\documentclass{ecai} 

\usepackage{multirow}
\usepackage{float}

\usepackage{latexsym}
\usepackage{amssymb}
\usepackage{amsmath}
\usepackage{amsthm}
\usepackage{booktabs}
\usepackage{enumitem}
\usepackage{graphicx}
\usepackage{color}
\usepackage[linesnumbered,ruled,vlined]{algorithm2e}

\newcommand{\BibTeX}{B\kern-.05em{\sc i\kern-.025em b}\kern-.08em\TeX}

\begin{document}


\begin{frontmatter}


\paperid{123} 


\title{Long-Short Distance Graph Neural Networks and Improved Curriculum
Learning for Emotion Recognition in Conversation}


\author[A]{\fnms{Xinran}~\snm{Li}}
\author[A]{\fnms{Xiujuan}~\snm{Xu}\thanks{Corresponding Author. Email: xjxu@dlut.edu.cn.}}
\author[A]{\fnms{Jiaqi}~\snm{Qiao}} 

\address[A]{School of Software Technology, Dalian University of Technology}


\begin{abstract}
Emotion Recognition in Conversation (ERC) is a practical and challenging task. This paper proposes a novel multimodal approach, the Long-Short Distance Graph Neural Network (LSDGNN). Based on the Directed Acyclic Graph (DAG), it constructs a long-distance graph neural network and a short-distance graph neural network to obtain multimodal features of distant and nearby utterances, respectively. To ensure that long- and short-distance features are as distinct as possible in representation while enabling mutual influence between the two modules, we employ a Differential Regularizer and incorporate a BiAffine Module to facilitate feature interaction. In addition, we propose an Improved Curriculum Learning (ICL) to address the challenge of data imbalance. By computing the similarity between different emotions to emphasize the shifts in similar emotions, we design a "weighted emotional shift" metric and develop a difficulty measurer, enabling a training process that prioritizes learning easy samples before harder ones. Experimental results on the IEMOCAP and MELD datasets demonstrate that our model outperforms existing benchmarks.
\end{abstract}

\end{frontmatter}


\section{Introduction}

Emotion Recognition in Conversation (ERC) \citep{hu2024recent} is a practical task with applications in areas like chatbots \citep{Lee2020StudyOE} and the analysis of opinions on social media \citep{chatterjee-etal-2019-semeval}. In ERC, the recognition of the target utterance's emotion considers the influence of other utterances, primarily from the speaker themselves and other participants. The speaker's emotions may change under the influence of other participants. Therefore, fully exploring the previous attitudes and emotions of participants and modeling the dialogue context is crucial to identify the emotion of the target utterance. Curriculum Learning \citep{CL} is a training strategy that simulates the human learning process by starting with simple samples and gradually increasing difficulty, aiming to improve the model's generalization ability and convergence speed. Current research mainly faces the following two issues:

(1) Most existing studies employ Graph Neural Networks (GNN) \citep{xu-etal-2024-escp} to model historical utterances, treating contextual utterances (past or future) as nodes in a graph, and encoding contextual features into the target utterance through graph aggregation mechanisms. However, current GNN-based methods tend to construct overly complex models, using different graphs to process short- and long-distance contextual features separately. Although this approach can extract rich contextual features, there is a high degree of similarity between short-distance and long-distance features, leading to feature redundancy. Additionally, redundant computations across different graphs not only increase computational complexity and resource consumption but may also hinder performance improvement.

(2) At present, most methods focus mainly on feature extraction or model architecture optimization, with relatively little attention given to improving the training process. These approaches often introduce excessively complex model designs and lengthy feature extraction procedures, resulting in increased computational costs while offering limited performance gains. Moreover, ERC datasets generally suffer from severe class imbalance, making it difficult for models to adequately learn the characteristics of low-frequency categories during training, thereby affecting overall performance and generalization ability \citep{HybridCL}. Therefore, optimizing training strategies to enhance model performance on imbalanced datasets remains a crucial challenge that needs to be addressed.

To address these challenges, we propose a multimodal ERC model, Long-Short Distance Graph Neural Network (LSDGNN), along with an Improved Curriculum Learning (ICL) strategy. LSDGNN leverages a Directed Acyclic Graph (DAG) \citep{DAG-ERC} to integrate short- and long-distance contextual features while using a Differential Regularizer to enhance feature diversity and mitigate redundancy. To enable features exchange between the long- and short-distance modules and enhance their features mutually, we introduce the BiAffine Module. These improve the model’s ability to capture subtle emotional nuances. ICL consists of a difficulty measurer and a training scheduler, which dynamically adjust training based on sample complexity \citep{curriculum-2024}. By assigning different transition weights to emotions based on similarity, the model focuses more on easily confused emotions, improving classification performance. Moreover, our experiments show that assigning higher difficulty weight to similar emotions yields better results than assigning higher difficulty weight to dissimilar emotions, which aligns with human intuition.

Experiments on IEMOCAP and MELD show that our model outperforms existing approaches, with particularly strong improvements on IEMOCAP.

Our contributions can be summarized as follows:

(1) We propose LSDGNN, a novel and effective multimodal ERC model that enhances emotion recognition through long-short distance DAG-based feature fusion, Differential Regularizer, and BiAffine Module.

(2) We introduce ICL, an improved curriculum learning training strategy based on Weighted Emotional Shifts based on similarity, which enhances emotion classification performance. This method is flexible, allowing for easy integration into any emotion recognition model to improve its generalization ability.

(3) Our model achieves state-of-the-art performance and will serve as a benchmark for future ERC research. The code and data have been released publicly on GitHub.\footnote{\url{https://github.com/LiXinran6/LSDGNN_ICL}}


\section{Related Work}

With the rapid development of natural language processing and human-computer interaction technologies, emotion recognition in dialogue (ERC) has gradually become a hot research topic. This section introduces the commonly used ERC methods and the application of curriculum learning in this field.

\subsection{Commonly Used Methods in the ERC Field}
In recent years, research in the ERC field has focused on three baseline models: Recurrent Neural Networks (RNN), Graph Convolutional Networks (GNN), and Transformer models.

In RNN-related research, ICON \citep{ICON} uses hierarchical modeling to capture global context and employs GRU to handle temporal dependencies between utterances; DialogueCRN \citep{DialogueCRN} introduces cognitive factors, enhancing the understanding of context at both the situational and speaker levels.

In GNN-based research, DialogueGCN \citep{DialogueGCN} treats dialogue utterances as graph vertices, with edges constructed based on context. MMGCN \citep{MMGCN} uses multimodal features as nodes, connecting three modalities within the same node while establishing connections among the same modality. DAG-ERC \citep{DAG-ERC} takes into account the identity of the speaker and location attributes when building a directed acyclic graph neural network. 

In transformer-related research, BERT-ERC \citep{BERT-ERC} improves the performance through suggestion texts, fine-grained classification modules and two-stage training, showing strong generalizability. 

With the rise of large language models (LLMs), methods such as InstructERC \citep{InstructERC}, BiosERC \citep{BiosERC}, and LaERC-S \citep{LaERC} have adopted generative architectures to address ERC tasks. However, considering the potential data leakage issues of LLMs on the MELD dataset, this paper compares the proposed method only with small-scale models.

Although these techniques have achieved great success, data scarcity remains the biggest problem faced by ERC \citep{emotrans}.

\subsection{Curriculum Learning}
Curriculum Learning (CL) \cite{CL}, a training strategy that simulates human learning processes, has recently been explored in the field of Emotion Recognition in Conversation (ERC). 

Yang et al. \citep{HybridCL} proposed the Hybrid Curriculum Learning framework, which focuses solely on textual features and combines conversation-level and utterance-level curriculum learning strategies. By using a difficulty measurer based on the frequency of "emotion shifts" and enhancing emotion similarity, this framework helps the model gradually learn complex emotional patterns.

Additionally, Nguyen et al. \citep{curriculum-2024} introduced the MultiDAG+CL method, which integrates Directed Acyclic Graphs (DAG) to combine textual, acoustic, and visual features within a unified framework, enhanced by Curriculum Learning to address challenges related to emotional shifts and data imbalance. 

These studies suggest that incorporating Curriculum Learning into ERC tasks aids models in better handling emotional variations and data imbalance, thereby improving the accuracy and robustness of emotion recognition.

\section{Methodology}
This section introduces the following five aspects. First, it presents the problem definition. Next, it discusses the multimodal feature extraction methods. Then, it explains the construction method of the simplified graph structure used in this paper. After that, it introduces the proposed Long-Short Distance Graph Neural Network. And finally, it presents the proposed Improved Curriculum Learning.

\subsection{Problem Definition}
In ERC, given a conversation consisting of a sequence of utterances, it is defined as $D = \{u_1, u_2, u_3,..., u_N\}$, where $N$ represents the total number of utterances. Each utterance is spoken by a single speaker, defined as $u_{i,s_j}$, indicating that the speaker of the $i_{th}$ utterance is $s_j$. The goal of ERC is to assign an emotion label $y_k \in Y$, such as joy or sadness, to each utterance $u_i$ in the conversation. $Y$ is a set of emotion labels.

The solution to this problem is to propose a function $f$ that takes an utterance $u_i$ as input and outputs the predicted emotion label $y_i$. The function $f$ should model the context only by considering past utterances $\{u_1, u_2, ..., u_{i-1}\}$ and not using future utterances $\{u_{i+1}, u_{i+2}, ..., u_N\}$.

\subsection{Feature Extraction}
We use modality-specific encoders to extract features. For the textual modality, RoBERTa \citep{RoBERTaAR} is used to extract features, while a Fully Connected Network (FCN) is employed for processing the acoustic and visual modalities, as shown in Equation \eqref{eq:feature}:
\begin{eqnarray}\label{eq:feature}
h_{i}^{a} = \text{FCN}_A(u_{i}^{a}), \quad h_{i}^{v} = \text{FCN}_V(u_{i}^{v}), \quad h_{i}^{t} = \text{RoBERTa}(u_{i}^{t})
\end{eqnarray}
where \( \text{FCN}_A \) and \( \text{FCN}_V \) represent Fully Connected Networks for the audio and visual modalities, respectively, and \( \text{RoBERTa} \) is the feature extractor for the textual modality. These encoders generate context-aware raw feature encodings \( h_{i}^{a}, h_{i}^{v}, h_{i}^{t} \).

For a given utterance \( u_i \) with available multimodal inputs, its multimodal feature vector shown in Equation \eqref{eq:concat}:
\begin{eqnarray}\label{eq:concat}
H_i^{0} = h_{i}^{a} \oplus h_{i}^{v} \oplus h_{i}^{t}
\end{eqnarray}
where \( \oplus \) denotes the feature concatenation operation.

\subsection{Constructing a Graph through Conversation}
Different researchers use various complex graphs to extract text features. However, overly complex ideas and methods often lead to redundancy in the extracted features. In this paper, we use a simple and efficient graph, the directed acyclic graph (DAG) .

DAG is represented as $G = (V, E, R)$, where the nodes, denoted by $V = \{u_1, u_2, ..., u_N\}$, correspond to the utterances in the conversation. The edges $(i, j, r_{ij}) \in E$ represent the feature propagation from $u_i$ to $u_j$, with $r_{ij} \in R$ indicating the type of relation associated with the edge. The set of relation types $R = \{0, 1\}$ includes two categories: type 1 indicates that the connected utterances are spoken by the same speaker, while type 0 indicates otherwise.

\begin{figure}[t]
  \centering 
  \includegraphics[width=0.45\textwidth]{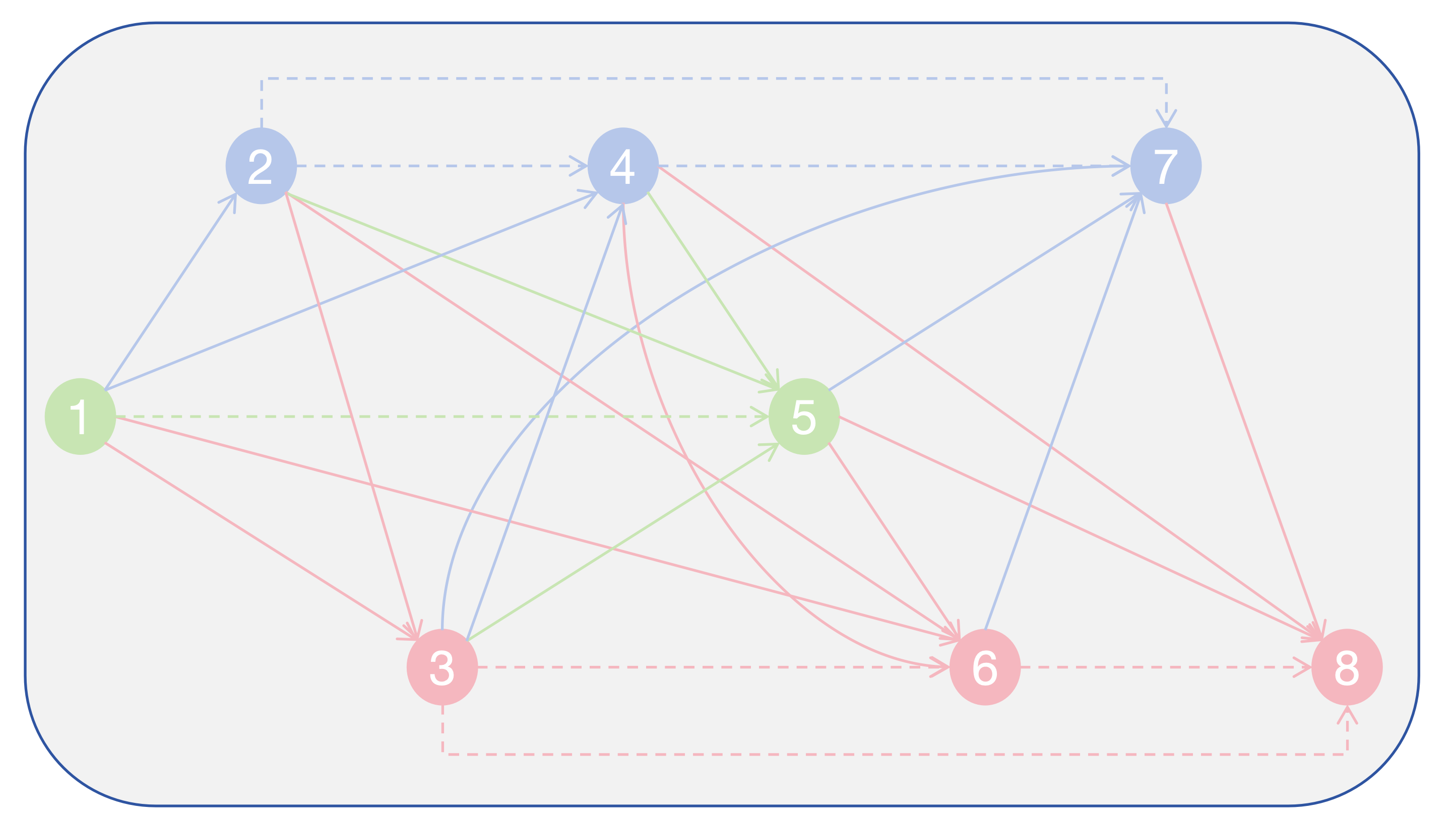} 
  \caption{A directed acyclic graph (DAG) constructed from a three-party conversation with the hyperparameter $\omega=2$. The utterances are ordered from left to right according to the speaking sequence. Dashed lines represent dependencies between the same speaker, while solid lines represent dependencies between different speakers. The speakers are represented in blue, green, and red.}
  \label{fig:node}
\end{figure}

The process of constructing the graph includes the following steps. First, each utterance in the conversation is treated as a node in the graph, and edges between nodes are determined based on the speaker’s identity. Starting from the second utterance, the algorithm sequentially checks previous utterances and adds edges with a relation type (1 or 0) depending on whether the speakers are the same. A maximum of $\omega$ utterances from the same speaker can be connected, and utterances from different speakers are also linked, ultimately generating a complete directed acyclic graph. The constructed graph is shown in the Figure \ref{fig:node}.

\subsection{Long-Short Distance Graph Neural Network}
\begin{figure*}[t]
  \centering 
\includegraphics[width=1\textwidth]{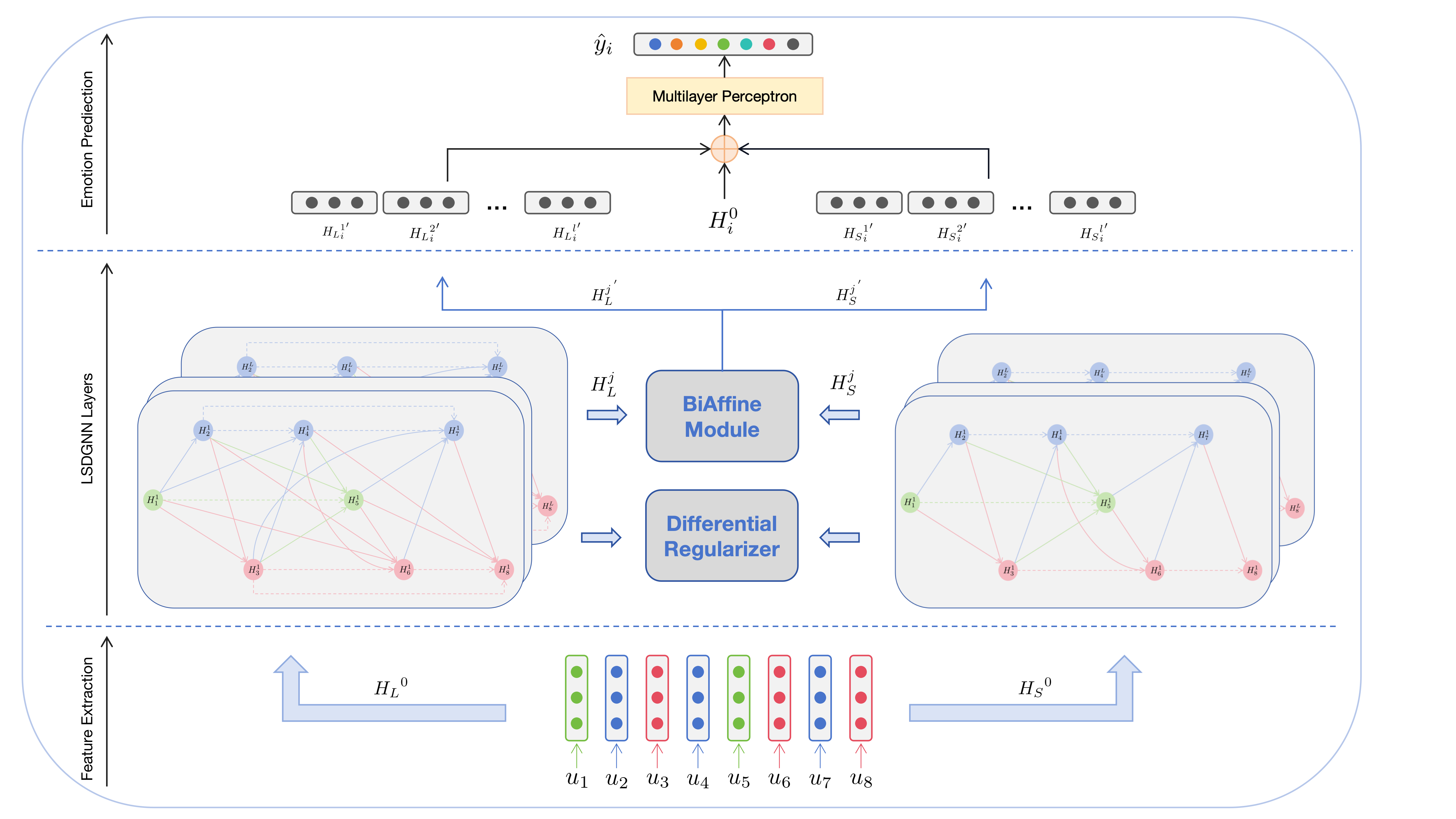} 
  \caption{The architecture diagram of LSDGNN. The left channel processes long-distance features, and the right channel processes short-distance features. The original inputs ${H_L}^0$ and ${H_S}^0$ are actually the same. In LSDGNN, at each layer, long-distance and short-distance features are processed using the Differential Regularizer and BiAffine Module. Here, $i$ represents the $i$-th utterance, and $j$ represents the features from the $j$-th layer.}
  \label{fig:structure}
\end{figure*}

In this subsection, we introduce the \textbf{L}ong-\textbf{S}hort \textbf{D}istance \textbf{G}raph \textbf{N}eural \textbf{N}etwork (LSDGNN) proposed in this paper. Based on DAGNN \citep{DAGNN}, this model constructs long-distance and short-distance DAGNNs separately to extract long- and short-distance features, respectively. Due to the redundancy between long- and short-distance features, we employ a Differential Regularizer to further ensure that these features are more distinct in representation. We also employ the BiAffine Module to facilitate feature interaction between the long- and short-distance modules, allowing their representations to mutually enhance each other. The framework of the model is shown in Figure \ref{fig:structure}.

\subsubsection{DAGNN}
DAGNN (Directed Acyclic Graph Neural Network) is a type of graph neural network designed to effectively handle directed acyclic graph data structures. It aggregates node features in temporal order, allowing the collection of features from multiple neighboring nodes at the same layer. Additionally, it permits feature propagation temporally within the same layer, enabling access to features from distant utterances. Each node's state update depends not only on its own state but also on the aggregated features from neighboring nodes. Thus, DAGNN effectively combines the advantages of graph neural networks (GNN) and recurrent neural networks (RNN). Its state update formula as shown in Equation \eqref{eq:dagnn}.
\begin{eqnarray}
    H^l_i = f\left(\text{Aggregate}\left(\{H^l_j \mid j \in N_i\}\right), H^{l-1}_i\right)
    \label{eq:dagnn}
\end{eqnarray}
where $H_i^l$ represents the representation of node $i$ in layer $l$. 

\subsubsection{Layers of LSDGNN}
The proposed LSDGNN is inspired by DAG-ERC \citep{DAG-ERC} and adopts its feature aggregation operation. Due to the temporal nature of the DAG, nodes must be updated sequentially from the first to the last utterance. For each utterance \( u_i \), the hidden state from the previous layer (\( l-1 \)) is used to compute attention weights on the hidden states of its predecessors at layer \( l \). The attention weights are computed as follows, as shown in Equation \eqref{eq:atten}:
\begin{eqnarray}
\alpha_{ij}^l = \text{Softmax}_{j \in N_i} \left( W_\alpha^l [ H_j^l \| H_i^{l-1} ] \right)
\label{eq:atten}
\end{eqnarray}
where \( W_\alpha^l \) are trainable parameters, and \( \| \) represents the concatenation operation. The model collects features through relation-aware transformations, enabling it to exploit different edge types. The aggregated features are then combined with the hidden state from the previous layer using a gated recurrent unit (GRU) to compute the current layer's hidden state. The aggregated features \( M_i^l \) are computed as shown in Equation \eqref{eq:agg}:
\begin{eqnarray}
M^l_i = \sum_{j \in N_i} \alpha_{ij} W^l_{r_{ij}} H^l_j
\label{eq:agg}
\end{eqnarray}
where \( W_{rij}^l \in \{ W_0^l, W_1^l \} \) are the learnable parameters for the relation-aware transformation. Independent learnable parameters are assigned to different edge types (type 1 or 0). Next, as shown in Equation \eqref{eq:gru_h}, the GRU processes the hidden state and aggregated features to compute \( \tilde{H}^l_i \):
\begin{eqnarray}
\tilde{H}^l_i = \text{GRU}^l_H(H^{l-1}_i, M^l_i)
\label{eq:gru_h}
\end{eqnarray}
where \( H_{i}^{l-1} \) and \( M_{i}^{l} \) are the inputs, and \( \tilde{H}_{i}^{l} \) is the output of the GRU. Additionally, a contextual feature unit is designed by swapping the inputs to the GRU, allowing the hidden state from the previous layer to control the propagation of contextual features. This process is described as shown in Equation \eqref{eq:gru_m}:
\begin{eqnarray}
C^l_i = \text{GRU}^l_M(M^l_i, H^{l-1}_i)
\label{eq:gru_m}
\end{eqnarray}
where \( C_{i}^{l} \) is the output of the contextual feature unit. Finally, the node’s representation at layer \( l \) is the sum of the outputs from the GRU and the contextual feature unit, as described in Equation \eqref{eq:final_rep}:
\begin{eqnarray}
H^l_i = \tilde{H}^l_i + C^l_i
\label{eq:final_rep}
\end{eqnarray}

\subsubsection{Feature Fusion and Prediction}
In a conversation, a speaker's emotion is influenced by previous utterances, which include both short-distance and long-distance utterances. Therefore, we design the LSDGNN model, which consists of two modules: one for long-distance and one for short-distance. For short distances, we set $\omega$ to 1, meaning the model traces back to the first self-uttered statement. Meanwhile, for long distances, we use a value greater than 1.

Through feature extraction, we obtain the multimodal fused feature $H^{0}$. Then, we input $H^{0}$ into the long-distance and short-distance channels, with the features denoted as $H_L^{0}$ and $H_S^{0}$, respectively. These two features are processed through multiple LSDGNN layers, and after each layer, the resulting features are denoted as $H_L^{j}$ and $H_S^{j}$, where $j$ indicates the $j$-th layer.

Inspired by Li et al. \citep{dual-graph}, we use mutual BiAffine transformations as a bridge to effectively exchange relevant features between the long-distance and short-distance modules. Specifically, the transformations are shown in Equations \eqref{eq:bi_affine_1} and \eqref{eq:bi_affine_2}.
\begin{eqnarray}
\label{eq:bi_affine_1}
    {H_{L}^{j}}^{'} = \text{softmax}\left( {H_{L}^{j}} W_1 ({H_{S}^{j}})^T \right) H_{S}^{j}
\end{eqnarray}
\begin{eqnarray}
\label{eq:bi_affine_2}
    {H_{S}^{j}}^{'} = \text{softmax}\left( {H_{S}^{j}} W_2 ({H_{L}^{j}})^T \right) H_{L}^{j} 
\end{eqnarray}
where $W_1$ and $W_2$ are trainable parameters.

For the $i$-th utterance, by concatenating the features ${H_{L}^{j}}^{'}$ and ${H_{S}^{j}}^{'}$ extracted from each layer through the BiAffine module with the original features $H^0$, the final feature representation $H$ is obtained as shown in Equation \eqref{eq:final_feature}.
\begin{eqnarray}
\label{eq:final_feature}
H = concat({H_{L}^{1}}^{'}, ..., {H_{L}^{j}}^{'}, ..., {H_{L}^{l}}^{'}, {H_{S}^{1}}^{'}, ...,  {H_{S}^{j}}^{'}, ..., {H_{S}^{l}}^{'}, H^0)
\end{eqnarray}
where $j$ ranges from 1 to $l$, indicating the features of the $j$-th layer. $l$ represents the total number of layers in the LSDGNN.

The obtained features $H_i$ of the utterance $u_i$ are fed through a feedforward neural network to predict the emotion, as shown in Equations \eqref{eq:emotion_relu}, \eqref{eq:emotion_softmax}, and \eqref{eq:emotion_argmax}.
\begin{eqnarray}
\label{eq:emotion_relu}
z_i = \text{ReLU}(W_H H_i + b_H)
\end{eqnarray}
\begin{eqnarray}
\label{eq:emotion_softmax}
P_i = \text{Softmax}(W_z z_i + b_z)
\end{eqnarray}
\begin{eqnarray}
\label{eq:emotion_argmax}
\hat{y}_i = \text{Argmax}_{k \in S}(P_i[k])
\end{eqnarray}

\subsubsection{Regularizer}
We expect the feature representations learned from the long-distance and short-distance modules to capture distinct characteristics. Therefore, we introduce a Differential Regularizer \citep{dual-graph} between the adjacency matrices of the two modules, as shown in Equation \eqref{eq:diff_reg}.
\begin{eqnarray}
\label{eq:diff_reg}
R_D = \frac{1}{\| A^{short} - A^{long} \|_F}
\end{eqnarray}

\subsubsection{Loss Function}
Our training objective is to minimize the following total objective function shown in Equation \eqref{eq:total_loss}:
\begin{eqnarray}
\label{eq:total_loss}
L = L_C + \lambda L_R
\end{eqnarray}
where $L_C$ is the standard cross-entropy loss, and $L_R$ is the differential regularization loss. $\lambda$ is the regularization coefficient, set to 0.1 here.

In ERC, the cross-entropy loss is formulated as shown in Equation \eqref{eq:loss}:
\begin{eqnarray}
\label{eq:loss}
L(\theta) = -\sum_{i=1}^{M} \sum_{t=1}^{N_i} \log P_{i,t}[y_{i,t}]
\end{eqnarray}
where \( M \) is the total number of samples in the training dataset. \( N_i \) denotes the number of time steps for the \( i \)-th sample. \( P_{i,t}[y_{i,t}] \) is the predicted probability for the true label \( y_{i,t} \) at time step \( t \) for the \( i \)-th sample.

\subsection{Improved Curriculum Learning}
Curriculum Learning \citep{CL} is a training strategy that arranges training tasks in a progressively increasing order of difficulty, improving the model's learning efficiency and performance. Moreover, it has excellent scalability and can be directly transferred to other models for training and usage.

\subsubsection{Difficulty Measure Function}
 Inspired by Yang et al. \citep{HybridCL}, we designed a Difficulty Measure Function based on the weighted emotional shift frequency in conversations, considering the emotional similarity between utterances. Our approach is simpler and more efficient.

As shown in Figure \ref{fig:circle}, we introduce a 2D arousal-valence emotion wheel, where each emotion label corresponds to a point on the unit circle. We calculate the similarity between emotion labels according to Equation \eqref{eq:similar}:

\begin{figure}[t]
  \centering 
  \includegraphics[width=0.45\textwidth]{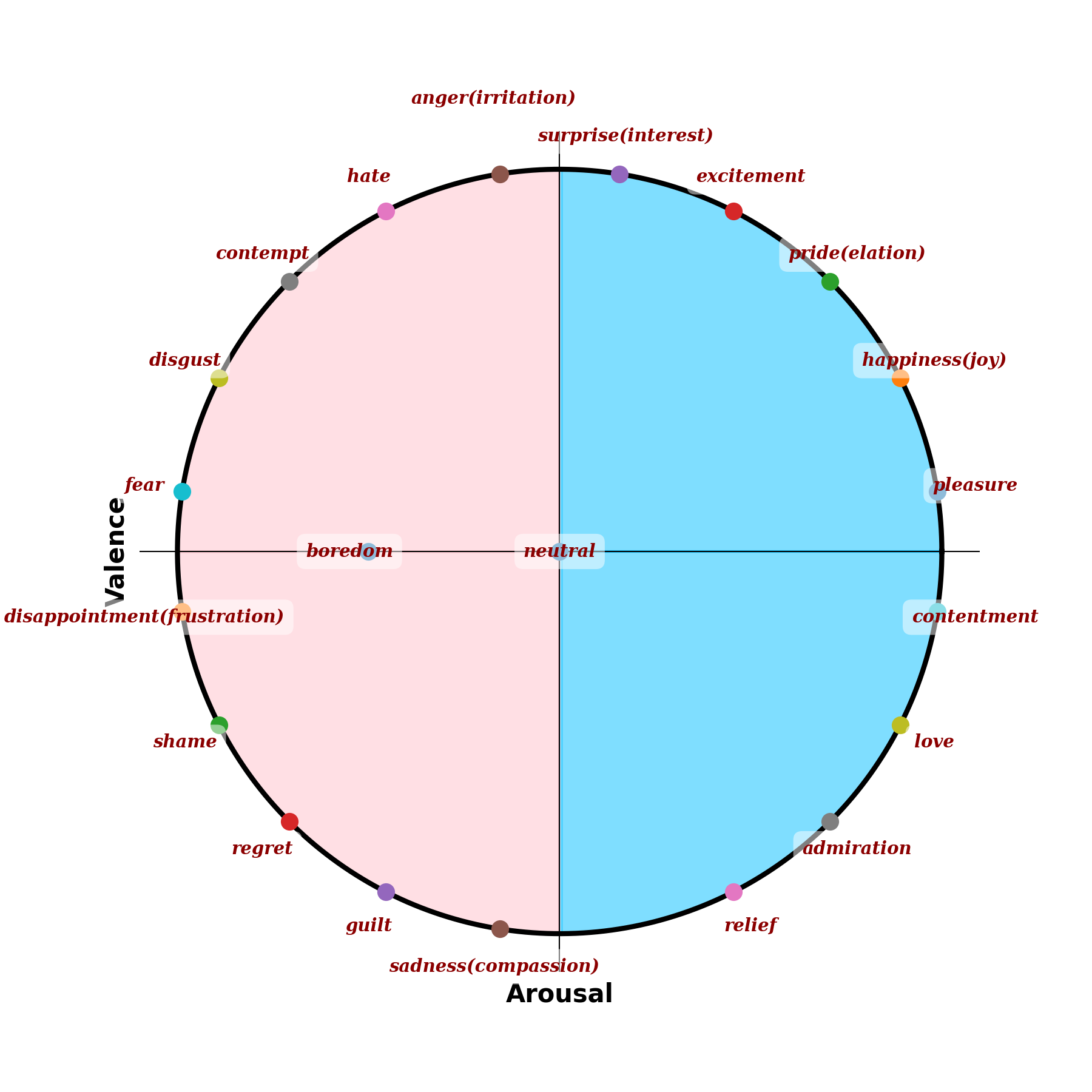} 
  \caption{Previous psychological research \citep{longago} suggested that emotions consist of two dimensions: Valence and Arousal, and emotions are described using a two-dimensional coordinate system similar to a wheel. Inspired by Jing et al. \citep{circle}, we construct this diagram, which includes all the emotions from the standard ERC dataset. Each emotion label can be mapped to a point on the unit circle.}
  \label{fig:circle}
\end{figure}

\begin{eqnarray}
s_{ij} =
\begin{cases}
\max(\cos(\theta_{ij}), 0) & \text{if } \mathbf{v}_i \cdot \mathbf{v}_j > 0 \\
0 & \text{if } \mathbf{v}_i \cdot \mathbf{v}_j < 0 \\
\frac{1}{N} & \text{if } \mathbf{v}_i \cdot \mathbf{v}_j = 0
\end{cases}
\label{eq:similar}
\end{eqnarray}
where $s_{ij}$ represents the similarity between label $i$ and label $j$, $\mathbf{v}_i$ denotes the valence value of label $i$, $\theta_{ij}$ is the angle between label $i$ and label $j$. $N$ is the total number of emotions in the dataset. The closer two emotions are, the higher their similarity value.

We have constructed a function to calculate the difficulty of a conversation based on the Weighted Emotional Shifts (WES). An emotional shift is defined as occurring when the emotions expressed in two consecutive utterances by the same speaker are different. A weighted emotional shift is defined as shown in Equation \eqref{eq:simi}:
\begin{eqnarray}
\label{eq:simi}
N^{WES} = k \times similarity + b
\end{eqnarray}

We employed a linear transformation, where the term $similarity$ refers to the degree of similarity between two emotions, $k$ represents the weight and $b$ denotes the bias. Thus, the difficulty of a conversation $c_i$ is defined as shown in Equation \eqref{eq:suanfa}:
\begin{eqnarray}
DIF(c_i) = \frac{\sum_{j=1}^{N_{shift}(c_i)}N_j^{WES}
 + N_{sp}(c_i)}{N_{u}(c_i) + N_{sp}(c_i)}
\label{eq:suanfa}
\end{eqnarray}
where ${N_{shift}(c_i)}$ and $N_{u}(c_i)$ represent the number of emotional shifts in conversation $c_i$ and the total number of utterances in conversation $c_i$, respectively. $N_{sp}(c_i)$ is the number of speakers appearing in conversation $c_i$, acting as a smoothing factor. $N_j^{WES}$ is the Weighted Emotional Shifts at the $j$-th emotional shift. The proposed algorithm is presented as Algorithm \ref{alg:training}. 

\begin{algorithm}[htbp]
\caption{Training with DMF based on Weighted Emotional Shifts (WES)}
\label{alg:training}
\KwIn{$D$ - training dataset, $M$ - training model, $k$ - number of buckets in the training scheduler, $DIF$ - difficulty measure function, $t$ - number of epochs, $n$ - number of utterances, $e$ - the emotion label of the utterances, $p(u_i)$ - the speaker’s corresponding utterance $u_i$, $S$ - Set containing the emotion sequence of speakers, $S[p[i]]$ - the emotion in the $i$-th utterance of speaker $p$, $WES$ - the total weighted emotional shifts of speak $p$ in a conversation.}
\KwOut{$M^*$ - the optimal model}

$S \gets \emptyset$, $N^{WES} \gets 0$, $N_{sp} \gets 0$, $N_u \gets 0$, $WES \gets 0$ \\
\For{$i = 1$ \textbf{to} $n$}{
    $S[p[i]] \gets S[p[i]] \cup \{e[i]\}$ \\
    $N_u \gets N_u + 1$
}
$N_{sp} \gets \text{number of unique speakers in the conversation}$ \\
\For{$p \in S$}{
    \For{$i = 1$ \textbf{to} $\text{length}(S[p]) - 1$}{
        \If{$S[p[i]] \neq S[p[i+1]]$}{
            $similarity \gets \text{get\_similarity}(S[p[i]], S[p[i+1]])$ \\
            $N^{WES} \gets k \times similarity + b$ \\
            $WES \gets WES + N^{WES}$
        }
    }
}

$DIF \gets \frac{WES + N_{sp}}{N_u + N_{sp}}$ \\
$D' \gets \text{sort}(D, DIF)$ \\
$D' \gets \{D_1, D_2, ..., D_k\}$ where $DIF(d_a) < DIF(d_b), \forall i < j, d_a \in D^i, d_b \in D^j$ \\
\For{$epoch = 1$ \textbf{to} $t$}{
    \If{$epoch \leq k$}{
        $D^{train} \gets D^{train} \cup D^i$
    }
    $\text{TRAIN}(M, D^{train})$
}

\Return{$M^*$}
\end{algorithm}


The time complexity of the algorithm is \(O(n + m \log m + t)\), where \(n\) is the number of utterances, \(m\) is the number of conversations, and \(t\) is the number of training epochs. When \(n\) is large, the complexity is dominated by \(O(n)\). The space complexity is \(O(mn)\), with dataset storage being the main factor.

\subsubsection{Training Scheduler}
The training scheduler organizes the training process by dividing the dataset $D$ into multiple bins ${\{D_1, D_2, ..., D_k\}}$ based on similar difficulty. Training starts with the easiest bin, and after several epochs, the next bin is gradually added. Once all bins have been used, additional training epochs are performed.

\section{Experimental Settings}
This section introduces the datasets and implementation details.

\subsection{Datasets}
We evaluated our approach on two ERC datasets: IEMOCAP \citep{IEMOCAP} and MELD \citep{MELD}. The detailed dataset statistics are shown in Table \ref{dataset}. We compared our method with several state-of-the-art baselines, including both unimodal and multimodal approaches. Consistent with previous work, the evaluation metric of our model is based on the average weighted F1 score from five random runs on the test set. Additionally, we include accuracy and macro F1 for supplementary comparison.

\begin{table}[H]
\caption{Statistics of the two datasets}
\centering
\small
\begin{tabular}{cccc}
\toprule
Dataset & Partition &  Utterance & Dialogues \\
\midrule
\multirow{2}{*}{IEMOCAP} & train + val & 5810 & 120 \\
& test & 1623 & 31 \\
\multirow{2}{*}{MELD} & train + val & 11098 & 1152 \\
& test & 2610 & 280 \\
\bottomrule
\end{tabular}
\label{dataset}
\end{table}

\subsection{Implementation Details}
 Each training and testing process runs on a single RTX 3090 GPU. For IEMOCAP, we set the number of LSDGNN layers to 4, dropout rate to 0.4, batch size to 16, embedding layer size to 2948, differential regularization loss to 0.1, the number of buckets in curriculum learning to 5, and the learning rate to 0.0005, with each epoch taking a maximum of 10 seconds. For MELD, we set the number of LSDGNN layers to 2, dropout rate to 0.1, batch size to 64, embedding layer size to 1666, differential regularization loss to 0.1, the number of buckets in curriculum learning to 12, and the learning rate to 0.00001, with each epoch taking a maximum of 8 seconds.

\begin{table*}[!t]
 \caption{Experimental results on IEMOCAP dataset}
\centering
\resizebox{\textwidth}{!}{%
    \begin{tabular}{lcccccccccc}
    \hline
     Model & Happy & Sad & Neutral & Angry & Excited & Frustrated & Accuracy & Macro-F1 & Wa-F1 \\ \hline
    DialogueRNN \citep{Dialoguernn} & 33.18 & 78.80 & 59.21 & 65.28 & 71.86 & 58.91 & 63.40 & - & 62.75 \\
     ICON \citep{ICON} &32.80 &74.40 &60.60 &68.20 &68.40 &66.20 &64.00 &- & 63.50 \\
     DialogueGCN \citep{DialogueGCN} & 42.75 & \textbf{84.54} & 63.54 & 64.19 & 63.08 & 66.99 & 65.25 & - & 64.18 \\
     MMGCN \citep{MMGCN} & 42.34 & 78.67 & 61.73 & 69.00 & \textbf{74.33} & 62.32 & - & - & 66.22 \\
     DAG-ERC  \citep{DAG-ERC} &47.59 &79.83 &69.36 &66.67 &66.79 &68.66 &67.53
&- & 68.03 \\
     LR-GCN \citep{LR-GCN}& 55.50 & 79.10 & 63.80 & 69.00 & 74.00 & \textbf{68.90} & 68.50 & - & 68.30 \\
     MultiDAG+CL \cite{curriculum-2024} &45.26 &81.40 &69.53 &70.33 &71.61 &66.94 &69.11 & - &69.08 \\
     CBERL \citep{CBERL} & \textbf{67.34} & 72.84 & 60.75 & \textbf{73.51} & 70.77 & 71.19 & 69.36 & - & 69.27 \\
     DER-GCN \citep{DER-GCN} & 58.80 & 79.80 & 61.50 & 72.10 & 73.30 & 67.80 & 69.70 & -  & 69.40 \\
     
     \textbf{LSDGNN+ICL} &47.20 &81.53 &\textbf{71.00} &69.74 &73.19 &68.58  &\textbf{70.35} & \textbf{68.54} &\textbf{70.24} \\
     \hline
    \end{tabular}}
    \label{tb:IEMOCAP}
\end{table*}

\begin{table*}[!t]
 \caption{Experimental results on MELD dataset}
\centering
\resizebox{\textwidth}{!}{%
    \begin{tabular}{lccccccccccc}
    \hline
     Model & Neutral & Surprise & Fear  & Sadness & Joy   & Disgust & Anger & Accuracy & Macro-F1 & Wa-F1 \\ \hline
      ICON \citep{ICON}  & -       & -        & -     & -       & -     & -       & -     & -     & -    & 56.30\\
    DialogueRNN \citep{Dialoguernn} & 73.50   & 49.40    & 1.20  & 23.80   & 50.70 & 1.70    & 41.50 & -     & -   & 57.03 \\
     DialogueGCN \citep{DialogueGCN} & -       & -        & -     & -       & -     & -       & -     & -     & -    & 58.10 \\
     MMGCN \citep{MMGCN}  & -       & -        & -     & -       & -     & -       & -     & -     & -    & 58.65 \\
     DAG-ERC  \citep{DAG-ERC} & 77.26       & \textbf{57.95}        &\textbf{27.87}     & 37.78       & 60.48     & \textbf{30.36}       & 49.59     & 63.98     & -    & 63.63 \\
    
     MultiDAG+CL \cite{curriculum-2024} & -       & -        & -     & -       & -     & -       & -     & -     & -    & 64.00\\
     \textbf{LSDGNN+ICL} & \textbf{77.58}       & 57.90        & 18.18     & \textbf{37.94}       & \textbf{61.70}     & 29.70       & \textbf{51.88}     & \textbf{64.67}     & \textbf{47.84}    & \textbf{64.07}\\
     \hline
    \end{tabular}}
    \label{tb:MELD}
\end{table*}

\section{Results and Analysis}
This section presents the comparison between our model and state-of-the-art methods, the results of ablation studies, and the impact of different values of long-distance $w$, as well as different values of $k$ and $b$ in Weighted Emotional Shifts on model performance.

\subsection{Comparison with the State of the Art}
Table \ref{tb:IEMOCAP} and Table \ref{tb:MELD} present the performance of our model on the IEMOCAP and MELD datasets, respectively, with bold values indicating the best results among all models. We compare the models based on three evaluation metrics: weighted F1, accuracy, and macro F1. The results of other models are taken from their original papers, with “-” indicating missing values. Our experiments are conducted using five random seeds, and the reported results are the averaged values.

Our approach, LSDGNN+ICL (which integrates the LSDGNN model with the Improved Curriculum Learning strategy), achieves SOTA performance on both the IEMOCAP and MELD datasets, outperforming previous methods by 0.84\% on IEMOCAP and 0.07\% on MELD.

\subsection{Ablation Experiments}
To investigate the importance and necessity of each module in our model, we conducted ablation experiments on both datasets. The results are presented in Table \ref{tab:ablation}, with each value averaged over five runs. All comparison metrics in this section are weighted F1 scores.
\begin{table}[H]
    \centering
    \caption{Ablation study results on the IEMOCAP and MELD datasets}
    \begin{tabular}{lcc}
        \toprule
        Model & IEMOCAP & MELD \\
        \midrule
        \textbf{LSDGNN+ICL} & \textbf{70.24} & \textbf{64.07} \\
        w/o ICL & 68.92 (↓ 1.32) & 63.78 (↓ 0.29) \\
        w/o ICL + DR & 68.70 (↓ 1.54) & 63.74 (↓ 0.33) \\
        w/o ICL + BM + DR & 68.55 (↓ 1.69) & 63.72 (↓ 0.35) \\
        w/o ICL + Long Distance & 68.08 (↓ 2.56) & 63.63 (↓ 0.44) \\
        \bottomrule
    \end{tabular}
    \label{tab:ablation}
\end{table}

From the results, it can be observed that removing any part of the model leads to a decline in performance, indicating that each module plays a significant role. When the Improved Curriculum Learning is removed, the performance on both datasets significantly drops. Similarly, when the long-distance module is removed, the performance also notably decreases. It can be seen that the Differential Regularizer and BiAffine Module contribute to improving the performance on long-distance tasks to some extent. When both ICL and the long-distance module are removed, the model essentially becomes equivalent to DAG-ERC.

\subsection{Different Values $w$ of Long-Distance}

$\omega$ refers to the number of previous nodes that have the same speaker as the current node when constructing the DAG. Specifically, the current node searches for $\omega$ nodes with the same speaker in the graph. Generally, a smaller $\omega$ searches for fewer previous utterances, while a larger $\omega$ searches for more previous utterances, connecting with a greater number of speech sentences. Our model consists of two modules: the long-distance module and the short-distance module. For the short-distance module, we default $\omega$ to 1, while different values of $\omega$ in the long-distance module have an impact on performance. Table \ref{tb:w} shows the effect of different $\omega$ values in the long-distance module on the model's performance. All comparison metrics in this section are weighted F1 scores, averaged over five random seeds.

\begin{table}[H]
\caption{Comparison of the effects of different $\omega$ values of Long-Distance on two datasets}
\centering
\resizebox{0.4\textwidth}{!}{%
\begin{tabular}{ccc}
\hline
Different Values $w$  & IEMOCAP & MELD \\ \hline
2 & 68.13	 & 63.66 \\ 
3 & 68.36 &63.51\\ 
4 & 68.58 &63.71\\ 
5 & \textbf{68.92}   &\textbf{63.78}\\
6 & 68.5	&63.62\\
$\infty$ &68.39 &63.65\\
\hline
\end{tabular}}
\label{tb:w}
\end{table}

It can be observed that as the $\omega$ value of the long-distance module increases from small to large, the model's performance on both datasets gradually improves, reaching its best performance when $\omega$ is 5. After that, the performance starts to decline slightly. We also tested the case where the $\omega$ of the long-distance module is set to infinity, and the result showed a slight decrease in performance, not reaching the best. This indicates that excessively long historical context does not further improve the model's performance, but may instead introduce too much redundant and irrelevant context, negatively affecting the model's effectiveness.

\subsection{The impact of parameters on Weighted Emotional Shifts}
From Equation \ref{eq:simi}, we can see that after obtaining the similarity between emotional shifts, we apply a linear transformation, assigning different values to $N^{WES}$ by controlling the slope $k$ and the bias $b$. A larger $N^{WES}$ indicates a higher difficulty in emotional shifts. Therefore, when $k$ is positive, greater emotional similarity results in higher overall difficulty, leading the model to first learn samples with larger emotional differences before learning those with similar emotions. Conversely, when $k$ is negative, greater emotional differences correspond to higher overall difficulty, causing the model to first learn samples with similar emotions before moving on to those with greater differences. As shown in Table \ref{tb:kb}, we experiment with different values of $k$ and $b$ and ultimately determine the optimal parameter combination for both datasets. All comparison metrics in this section are weighted F1 scores, averaged over five random seeds.

\begin{table}[H]
\caption{The impact of different $k$ and $b$ values on model performance}
\centering
\resizebox{0.4\textwidth}{!}{%
\begin{tabular}{cc|cc}
\hline
values of $k$ &values of $b$  & IEMOCAP & MELD \\ \hline
\multirow{6}{*}{1}  &0 & 69.03	 & 63.83 \\ 
                    &0.1  & 69.4 &\textbf{64.07}\\ 
                    &0.2  & 69.14 &64.06\\ 
                    &0.3  & 69.31   &63.84\\
                    &0.4  & \textbf{70.24}	&63.91\\
                    &0.5  & 69.67	&63.9\\\hline
\multirow{6}{*}{-1} &1    & 70.0	 &64.05\\
                    &1.1 & 69.27 &64.01\\
                    &1.2 & 69.09 &63.83\\
                    &1.3 & 69.21 &63.81\\
                    &1.4 & 68.64 &63.88\\
                    &1.5 & 68.49 &63.68\\
\hline
\end{tabular}}
\label{tb:kb}
\end{table}

We can see that when $k$ is positive, regardless of the value of $b$, the overall results are better than those when $k$ is negative. This indicates that curriculum learning has indeed played a role, while also proving that the more similar the changing emotions are, the harder they are to learn, which aligns with human intuition. In addition, different values of $b$ cause significant fluctuations in model performance, especially when $k$ is -1. This further suggests that when $k$ is negative, regardless of the value of $b$, the defined difficulty standard makes it harder for the model to learn, thereby weakening the effect of curriculum learning. When $k$ is 1, all results show varying degrees of improvement regardless of the value of $b$, whereas when $k$ is -1, some results exhibit a significant decline.

\section{Conclusion}
In this paper, we present a novel multimodal approach called the Long-Short Distance Graph Neural Network (LSDGNN) for Emotion Recognition in Conversation (ERC). LSDGNN combines both long- and short-distance graph neural networks based on a Directed Acyclic Graph (DAG) to extract multimodal features from distant and nearby utterances. To ensure effective representation of these features while maintaining mutual influence between the two modules, we introduce a Differential Regularizer and incorporate a BiAffine Module to enhance feature interaction. Additionally, we propose an Improved Curriculum Learning (ICL) approach to address the data imbalance issue by designing a "weighted emotional shift" metric, which emphasizes emotional shifts between similar emotions. This difficulty measurer enables the model to prioritize learning easier samples before more difficult ones. The method can also be directly transferred to other ERC tasks. Experimental results on the IEMOCAP and MELD datasets demonstrate that our model outperforms existing benchmarks, showcasing its effectiveness in addressing the complexities of ERC tasks.

In the future, we will explore more advanced transformation methods for weighted emotional shifts and investigate adaptive graph strategies for multi-party and low-resource scenarios.


\begin{ack}
This work is funded in part by the National Natural Science Foundation of China Project (No. 62372078).
\end{ack}


\bibliography{mybibfile}

\end{document}